\documentclass{article} % For LaTeX2e
\usepackage{iclr2016_workshop,times}
\usepackage{hyperref}
\usepackage{url}
\usepackage[pdftex]{graphicx}
\usepackage{longtable}
\usepackage{amsmath}

\title{Temporal Convolutional Neural Networks for Diagnosis from Lab Tests}

\author{Narges Razavian, David Sontag\\
Computer Science Department, New York University\\
New York City, NY \\
\texttt{\{razavian,dsontag\}@cs.nyu.edu} \\
}

\newcommand{\beginsupplement}{%
        \setcounter{table}{0}
        \renewcommand{\thetable}{S\arabic{table}}%
        \setcounter{figure}{0}
        \renewcommand{\thefigure}{S\arabic{figure}}%
}
\begin{document}
\maketitle
\begin{abstract}
Early diagnosis of treatable diseases is essential for improving healthcare, and many diseases' onsets are predictable from annual lab tests and their temporal trends. We introduce a multi-resolution convolutional neural network for early detection of multiple diseases from irregularly measured sparse lab values. Our novel architecture takes as input both an imputed version of the data and a binary observation matrix. For imputing the temporal sparse observations, we develop a flexible, fast to train method for differentiable multivariate kernel regression. Our experiments on data from 298K individuals over 8 years, 18 common lab measurements, and 171 diseases show that the temporal signatures learned via convolution are significantly more predictive than baselines commonly used for early disease diagnosis.
\end{abstract}

\section{Introduction}
Representation learning and unsupervised feature discovery via deep learning has led to ground breaking advances in domains such as image processing \citep{krizhevsky2012imagenet}, speech recognition \citep{graves2005framewise}, natural language processing \citep{mikolov2013distributed}, surpassing methods based on hand-engineered features in all benchmarks tested. Following recent availability of large electronic medical record datasets and other biological signals \citep{hsiao2014trends}, discovery of early temporal disease signatures within lab values has become a possibility. In this paper, we are interested in discovering these signatures to perform early diagnosis of multiple preventable and treatable diseases.

There are many challenges associated with performing machine learning on observational medical data. Data is almost never missing at random and often sparse. Labels such as disease onsets, if they exist, are noisy. Many unobserved variables also affect the outcome. Each individual has a different baseline healthy state, and variations compared to their own baseline indicates whether they have deviated from their optimal health state. This last characteristic has inspired us to train a temporal convolution model \citep{le1990handwritten,lecun1998gradient,tompson2014joint,krizhevsky2012imagenet} to learn variation patterns of labs as biological representations of healthy and diseased states. In the clinical domain, each biomarker varies with a different natural speed of change in the body. Therefore, in this paper we focus on multi-resolution deep convolutional architectures, inspired by \citet{mnih2014recurrent}.

%Temporal convolution on time series has a long history in machine learning \cite{} \cite{} \cite{}, but models that operate on sparse irregularly sampled multivariate time series are less studied. 
Strong biases exist in the frequency and timing of lab measurements. For instance, a person suspected to have diabetes is likely to have more Glucose lab tests ordered by the physician. As we will see later in the experiments, the {\em utilization} signals (i.e. how often and when each lab is ordered) are predictive of disease onset as well. But recent health care developments, such as Theranos lab testing startup or affordable wearables capable of measuring different chemicals at home, will likely result in the process of obtaining lab tests becoming both significantly cheaper and easier. As a result, we expect that this utilization signal will look very different in a few years from how it is today. Additionally, medical community has actively studied the causal effect of variations on different signals, such as glucose \citep{kilpatrick2007relating}, cholesterol \citep{bangalore2015visit}, blood pressure\citep{hata2013effects}, and prostate-specific antigen \citep{roehrborn1996variability} on different disease onsets. A model trained on the imputed biological measurements rather than the healthcare utilization signals can better aid with hypothesis generation for such causal studies, which can lead to meaningful interventions. For these reasons, and to provide models that reveal the disease signatures (i.e. changes in the actual chemicals in the body prior to a disease onset), we consider models that work on an imputed version of the lab data.

% In this paper, we explore architectures which both use the underlying raw data and a combination of the observation pattern and an imputed dataset. % important to ensure that disease signatures are only based on biological signals rather than the utilization. 

A generative model that captures all sources of bias could potentially remove the utilization effects using marginalization and inference. Examples of such models include Gaussian processes and generative models based on recurrent neural networks \citep{sutskever2011generating,sutskever2009recurrent,tang2014learning,chung2015recurrent}. However, for high dimensional structured continuous input time series, when the variables are not observed at the same time, marginalization can be prohibitively slow or not possible. %An alternative solution is to perform imputation first to separate the biological signal from healthcare utilization signal, and then to learn disease signatures on the imputed data. 

Instead, we propose a convolution based formulation of multivariate nonparametric (kernel) regression, which is capable of inferring the structure of the input as part of the imputation task. This approach can obtain competitive results as Gaussian processes for univariate data, and is extremely fast to train for asynchronous multivariate data. Moreover, although we do not explore it in this paper, this approach to imputation is amenable to end-to-end training together with the supervised prediction task.

Our paper is structured as follows: We first present our prediction model architecture, which is a multi-resolution convolution network with shared components for multi-task learning. We then present our imputation model architecture, which is based on a differentiable formulation of nonparametric (kernel) density estimation, for single time-series as well as multiple dependent time-series. The final architecture is the combination of the imputation network with the prediction network. Our evaluation is performed on an original dataset of 298,000 individuals tracked for 8 years. We use temporal observations of 18 most commonly measured lab measurements and perform early (at least 3 months in advance) detection of 171 diseases and conditions. We compare our imputation and prediction results to an extensive set of baselines and various input signals, and show that the temporal signatures learned via convolution are significantly more predictive than baselines commonly used for early disease diagnosis.

Although we present the new multi-resolution deep convolutional architecture and the multivariate nonparametric regression algorithms in the context of early diagnosis from lab tests, we emphasize that both of these algorithms are much more broadly applicable to prediction problems in machine learning with temporal, sparse and irregularly measured, multivariate data. User behavior modeling, financial data analysis, and providing useful service from wearables are among domains where the data exhibits similar characteristics and challenges.
\section{Temporal Convolutional Network}
We formulate the task of diagnosis as a supervised multi-task classification task. Each individual has a variable-length history of lab observations ($X$) and diagnosis records ($Y$). $X$ is continuous valued, and $Y$ is binary. We use a sliding window framework to deal with variable length input. At each time point $t$ for each person $X$, the model looks at a backward window of 36 months of all $D$ biomarkers of the input, $X_{t-36:t}^{1:D}$, to predict the output. Output is a binary vector $Y$ of size $M$, corresponding to $M$ disease onsets each happening within the following months from $t+3$ to $t+3+24$. In this paper we consider $18$ commonly measured biomarkers, and predict $171$ common diseases (i.e. $D=18$, $M=171$). To retain clinical validity for our early detection task, we have carefully designed our experimental setting, outcome definition and exclusion criteria, which we discuss in details in section \ref{sec:res_pred}.

\begin{figure}[h]
\begin{center}
\includegraphics[width=1\linewidth]{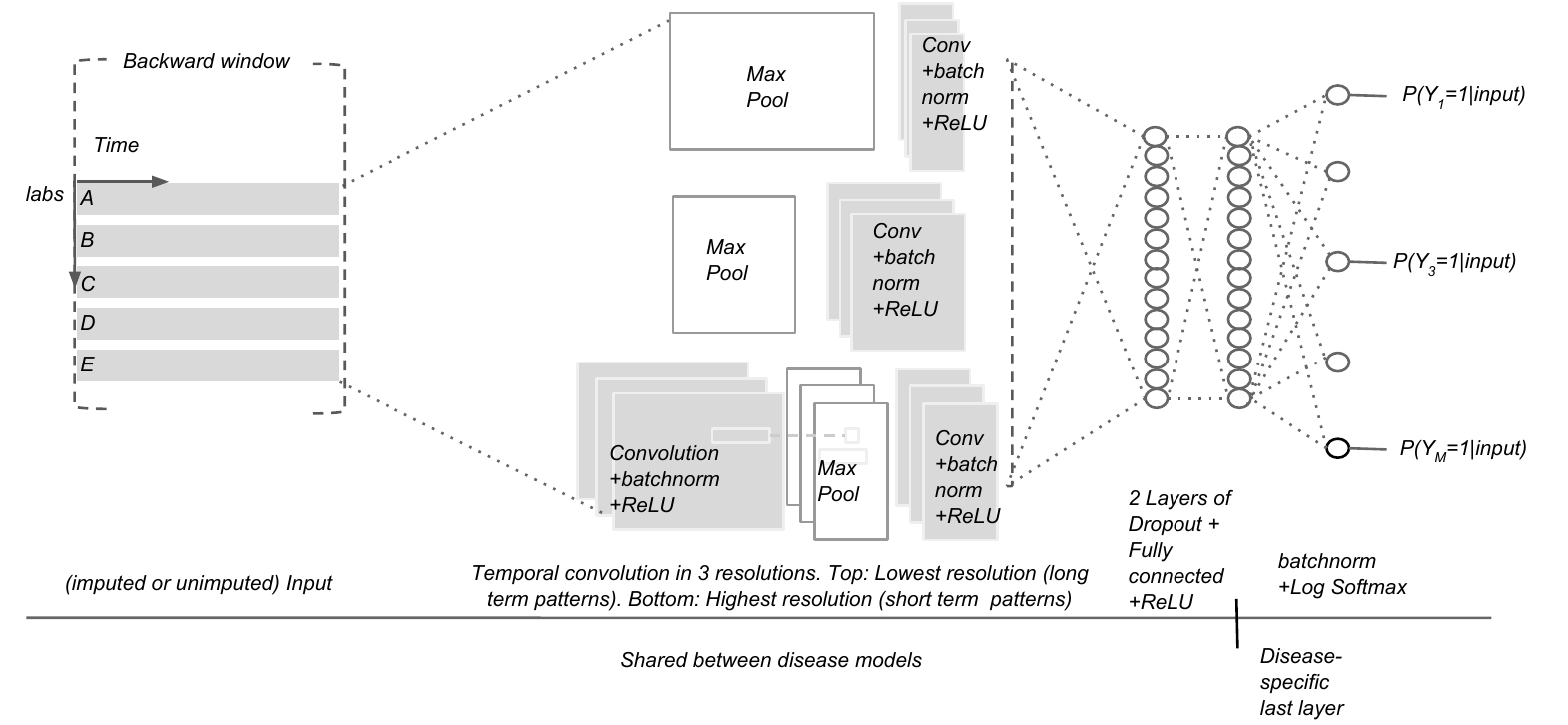}
\end{center}
\caption{Network architecture for input to output prediction.}\label{fig:predarch}
\end{figure}

Our temporal convolution model is shown in figure \ref{fig:predarch}. The input to the model can be raw(un-imputed) observations; imputed observations; or the concatenation of the imputed data and the binary observation pattern. The choice of the input will allow us to analyze the nature of signals that better predicts each disease. Binary observation pattern only encodes the health-care interaction signal, which is subject to fast change as the health-care policies and the economy change. While currently useful, these health-care interaction signals will have different distributions in the era where preventive medicine is in full practice. Therefore, a model which relies on the chemical state of the body (i.e. imputed observations) would be required. We present the imputation network in section 3.3 and the full model of imputation and prediction is shown in figure \ref{fig:fullarch}.

The prediction part of network is shown in figure \ref{fig:predarch}.
%, all diseases share the underlying temporal patterns, and only differentiate in the last layer. 
Specifically, we defined $X_{t-36:t}^{1:D}$ to be the input of the network at time $t$. Let there be $J$ number of filters (or patterns) in each of the levels of the multi-resolution convolution network. Each filter $K_{i}^j$ ($j=1:J$) is of size $1 \times L$, corresponding to temporal filters of size $L$ at different resolution levels ${i}$. The third level of resolution includes two layers of convolution corresponding to filters $K_{3}^j$ and $K_{5}^j$. The output of the multi-resolution convolution network is a vector $C = [C_1, C_2, C_5]$ which is defined as follows:

\begin{align}
 C_1^{d,j} =& f(b_1^j + (K_1^j * MaxPool(X_{t-36:t}^{d}, p^2) )) \\ 
 C_2^{d,j} =& f(b_2^j + (K_2^j * MaxPool(X_{t-36:t}^{d}, p) )) \\
 C_3^{d,j} =& f(b_3^j + (K_3^j * X_{t-36:t}^{1:D})) \\
 C_4^{d,j} =& MaxPool(C_3^{d,j}, p) \\
 C_5^{d,j} =& f(b_5^j + \sum_{k=1}^{J} K^j_5 * C_4^{d,k})
\end{align}

In the above definition, $*$ is a standard convolution operation. $f$ is a ReLU nonlinearity function \citep{nair2010rectified}. The vector $C_i$ is the concatenation of $C_i^{d,j}$ for all biomarkers $d=1:D$ and filters $j=1:J$, and $MaxPool(X,p)$ corresponds to non-overlapping max pooling operation defined as $MaxPool(Z,p)[i] = max(Z[ p\cdot i : p\cdot(i+1)-1])$ for each $i = 1:floor(length(Z)/p)$. The value of $p$ is set to 3 in our case. $b_i^j$ is a bias term and is learned during training. After every convolution operation we use batch normalization \citep{ioffe2015batch}. 

After the multi-resolution convolution is applied, the vector $C$ represents the application of filters to all biomarkers, and we note that the filters are shared across all biomarkers. We then use $2$ layers of hidden nodes to allow non-linear combination of filter activations on different biomarkers. 

\begin{align}
 h_1 = & f ( W_1^T C + b_{h_1}) \\
 h_2 = & f ( W_2^T h_1 + b_{h_2})
\end{align} 

$W_i$ is the weight of the hidden nodes and $b_{h_i}$ is the bias associated with each layer. Each of the hidden layers are subject to Dropout\citep{srivastava2014dropout} regularization (with probability 0.5) during training, and are followed by batch normalization. 

Finally, for each disease $m=1:M$, the model predicts the likelihood of the disease via logistic regression over $h_2$.

\begin{align}
P(Y_{m} = 1 | X_{t-36:t}^{1:D}) = \sigma(W_m^T h_2 + b_m)
\end{align}

The loss function for each disease is the negative log likelihood of the true label, weighted by the inverse label ratio to handle class imbalance during multi-task batch training. Diseases are trained independently, but the gradient is propagated through the shared part of the network.

Figure \ref{fig:predarch} is shown for imputed or un-imputed input. Prediction model for the case where input is the concatenation of imputed and binary observation mask is identical, except that the input is then of size $18\times2$ (i.e. vertical concatenation) times length of the backward window (36 months). Specific architectural choices and values of hyper-parameters are described in section \ref{sec:pred_spec}

%We pre-train the imputation model via mean squared error(MSE) optimization. In practice, our multivariate kernel regression is a form of hierarchical normalized convolution, where each variable learns the clique of variable it depends on, in the form of convolution filter and applies the filter. It then normalizes the results by the same filter applied only to the data observation binary mask. We now show the qualitative results and the form of the learned filters (or kernels).

\subsection{Related Work}
Medical field has been dominated by traditional feature engineering methods. Only recently, attempts to learn the patterns has started to gain some attention. \citep{lasko2013computational} studied a method based on sparse auto-encoders to learn temporal variation features from 30-day uric acid observations, to distinguish between gout and leukemia. \citep{che2015deep} developed a training which allows prior domain knowledge to regularize the deeper layers of feed-forward network, for the task of multiple disease classification when datasets are small. To our knowledge, a full scale study of convolutional neural networks for the task of disease pattern discovery has not yet been performed.

Within the domain of temporal convolutional networks, \citep{abdel2012applying,sainath2013deep} were among the first to show significant gains in speech recognition tasks in large scale. Unlike speech domain where the input is fully observed, in our case we have sparse and asynchronously measured observations. Alternative models would be recurrent neural network(RNN) models such as variants of LSTM, however given the state of the body in the past few years, there is no clear evidence that longer term dependencies are necessary. In addition, the trends on the biomarkers which are directly learned via temporal convolution might provide more clinically interpretable results currently. For these reasons we focus on temporal multi-resolution convolution model in this paper. 

% but due to the highly biased missing-ness patterns, a naive RNN would mix biological disease signatures and health-care interaction signals together. Our proposed model carefully separates the health-care interaction signals from the biological signal, and allows us to recover the predictive patterns easily, which would not be trivial using Recurrent representation learning models.

\section{Imputation via Differentiable Kernel Regression}\label{sec:kr}
In order to learn biological disease signatures, we now present our imputation model which we apply to the input prior to learning the variation patterns. Our model is based on nonparametric regression, which we formulate as differentiable functions of (univariate and multivariate) kernels and input. Using back-propagation \citep{rumelhart1988learning}, we then show how one can learn the entire form of the kernel function instead of cross validating within a limited set of parametric family (such as Gaussian or Laplace). We compare our method to Gaussian processes and traditional nonparametric(kernel) regression which use standard kernel functions.

\subsection{Related Work on Univariate Kernel Learning}
Within the field of nonparametric methods, most existing work only cross validate over a few well-known kernel functions such as Radial basis(Gaussian), Laplace, or other simple kernels, and fail to consider the entire space of legal kernels. In best case, attempts such as \citep{duvenaud2013structure}, \citep{gonen2011multiple} learn a composition or combination of kernel families. The algorithms are slow, and in practice, the search algorithm is not comprehensive enough to guarantee recovery of the correct kernel. Additionally learning multivariate kernels are also not possible using these methods. Our proposed solution overcomes all these issues.

\subsection{Univariate kernel regression: Learning the Kernel}
Imagine the input to be samples from D time series, each sampled irregularly. We denote the samples as ${x^1_{t^1_1}}$, ${x^1_{t^1_2}}$, ..., ${x^1_{t^1_{n_1}}}$, ...,${x^D_{t^D_1}}$, ${x^D_{t^D_2}}$, ..., ${x^D_{t^D_{n_D}}}$, where $x^d$ refers to time series $d$ and $t^d_1$,... $t^d_{n_d}$ refer to the time points over which time series $d$ is sampled. Kernel regression provides a general formalism for estimating any function with additive noise, provided that the signal is locally stationary.
Let's start from a single time series, $x(t)$. Kernel regression assumes the following:
$$ x = f(t) + \epsilon $$
$$\epsilon \sim N(0,\sigma^2)$$
Given observed samples $x_{t_1}$,...$x_{t_n}$ from the series, general function regression with additive noise lets us estimate the value of $x$ at a new time point $t_{new}$ as follows. 
$$x(t_{new}) = \mathbf{E}_{x \sim P(x|t=t_{new})}[x] $$
$$\mathbf{E}_{x \sim P(x|t=t_{new})}[x] = \int_x x P(x|t=t_{new}) dx =\int_x x \frac{P(x , t=t_{new})}{P(t_{new})} dx $$
At this point, one can use kernel density estimation to estimate the probabilities $P(x , t=t_{new})$ and $P(t_{new})$ from the training data. \citep{nadaraya1964estimating} and \citep{watson1964smooth} showed that using a positive semidefinite kernel function $K(t, t')$, the nonparametric regression formulation is reduced to:
\begin{equation}
\mathbf{E}_{x \sim P(x|t=t_{new})}[x] = \frac{\sum_{i=1}^n{x_{t_{i}}K(t_{new}, t_{i})}} {\sum_{i=1}^n{K(t_{new}, t_{i})}}
\end{equation}
We can now rewrite the nonparametric regression using convolution operator. To be able to use functional notation, we first write the sequence of observed samples {$x_{t_1}$,...,$x_{t_n}$} as a function: $\bar X_{train}(t)$ = $\sum_{i=1}^{n} x_{t_i}\delta(t,{t_i})$, where $\delta(t,\tau_0) = 1$ when $t=\tau_0$, and 0 otherwise. 

Denoting convolution operator as $\ast$, i.e. $(K \ast f)(t) = \int_\tau K(t-\tau) f(\tau) d\tau$, the numerator of the kernel regression is equal to:
$ \sum_{i=1}^n{x_{t_{i}}K(t_{i} - t_{new})} = (K \ast \bar X_{train}) (t_{new})$.
The denominator $P(t_{new})$ can similarly be written as a convolution of the kernel function with a sequence of $1$s at each point at which we have a sample, denoted as $I(\bar X_{train}:observed)(t) = \sum_{i=1}^{n}\delta(t,{t_i})$.
$$ \sum_{i=1}^n{K(t_{new}, t_{i})} = (K \ast I(\bar X_{train}:observed)) (t_{new}) $$
So the kernel regression formulation of Nadaraya and Watson reduces to the following formulation.
$$ \mathbf{E}_{x \sim P(x|t=t_{new})}[x] = \frac{(K \ast \bar X_{train}) (t_{new})}{(K \ast I(\bar X_{train} :observed)) (t_{new})} $$
This formulation has previously been used in image processing literature under the name {\em normalized convolution} \citep{knutsson1993normalized}, however only parametric kernels have been considered before.  By writing the kernel regression as a fully differentiable function, we can now learn $K(\tau)$ at each position $\tau$ within the kernel domain via back-propagation. We can also compose this differentiable kernel regression module within any subsequent differentiable operators and perform multiple tasks. 

\begin{figure}[h]
\begin{center}
\includegraphics[width=0.8\linewidth]{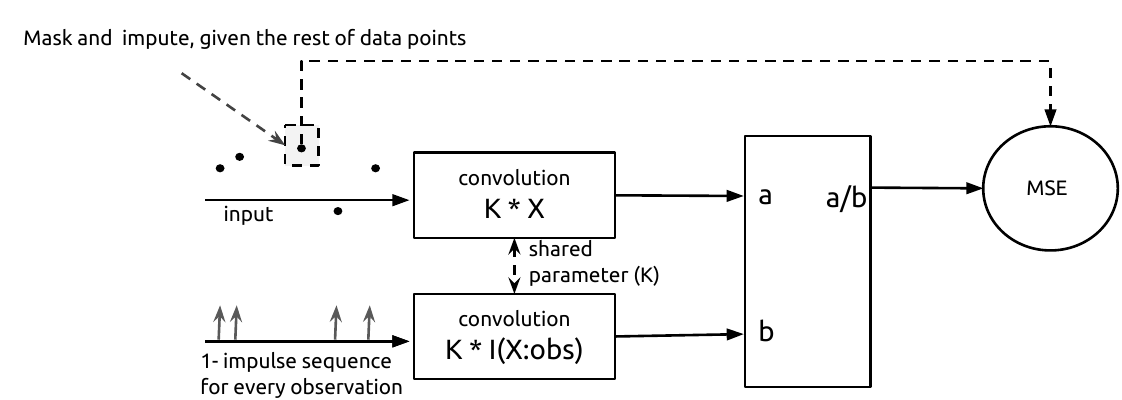}
\end{center}
\caption{Architecture for differentiable univariate kernel regression.}\label{fig:arch}
\end{figure}

In this paper we use leave-one-out imputation mean squared error as the loss function. In practice, we assume the domain of $K$ is bounded between $[-M, M]$, therefore the learning task will have $2M+1$ parameters. Figure \ref{fig:arch} shows the architecture of this model.

\subsection{Multivariate kernel regression: Learning the Temporal Kernel and Dependency structure}

Let's now assume that we have $D$ time series, corresponding to each of the labs. We could attempt to model the full joint distribution of the time series,
so that observations of related labs at nearby times could be used to infer the values of missing labs. Various multi-output extensions of Gaussian processes have been proposed previously. The dominant approaches rely on Bayesian formalization and process convolution \citep{boyle2004dependent,alvarez2010efficient,alvarez2009sparse}, and require known dependency structure on the multiple outputs. The main problem with the models is that they are only scalable under the sparse structure assumptions \citep{alvarez2011kernels,byron2009gaussian}. In biological domains, observed variables are highly correlated due to many unobserved latent variables. In the general high-dimensional tightly correlated setting, considering that the parameters of the kernels need to be tuned via cross-validation, these models are not scalable. One solution for unconstrained structure was proposed in \citep{wilson2012gaussian}, however inference required Monte Carlo sampling or variational inference, which were inefficient. Alternatively, one could model the full joint distribution for a window of interest using a nonparametric graphical model \citep{fukumizu2007kernel,smola2007hilbert,song2011kernel}. However, since the labs are measured asynchronously with significant missing data, inference and learning of these models can be extremely slow. Our proposed framework allows us to easily extend the univariate kernel regression to the multivariate setting, giving a very fast and -- as we show in the experiments -- accurate multivariate approach.

\subsection{Multivariate Kernel Regression}
We extend the kernel $K$ to be a matrix of size $D \times (2M+1)$, and learn the kernel magnitude at (r,j,s) corresponding to kernel value between the imputed series at time $r$ and series $j$ at time $s$. Multivariate kernel regression becomes a 2D convolution of this kernel matrix with all time series' observed points in the numerator, normalized by the 2D convolution of the kernel matrix with a binary matrix encoding which series at which time point does have a nonzero observation. 
$$ \mathbf{E}_{x^d \sim P(x^d|t=t_{new})}[x^d] = \frac{(K \ast \bar X^{1..d}_{train})(t_{new})}{(K \ast I(\bar X^{1..d}_{train} :observed)) (t_{new})} $$
% 
%This formulation allows every variable at time t to be a linear function of all other variables between $t-M$ and $t+M$. An example shown in the experiments section includes a lab value that is a ratio of two other lab values by definition. If we hadn't known this in advance and applied multivariate kernel regression, based on this formulation, the lab value is approximated with the difference of the two observations instead of the ratio. In practice, however, for many settings this model is sufficient and has strong empirical performance as we will show in experiments section.

\begin{figure}
\begin{center}
\includegraphics[width=0.8\linewidth]{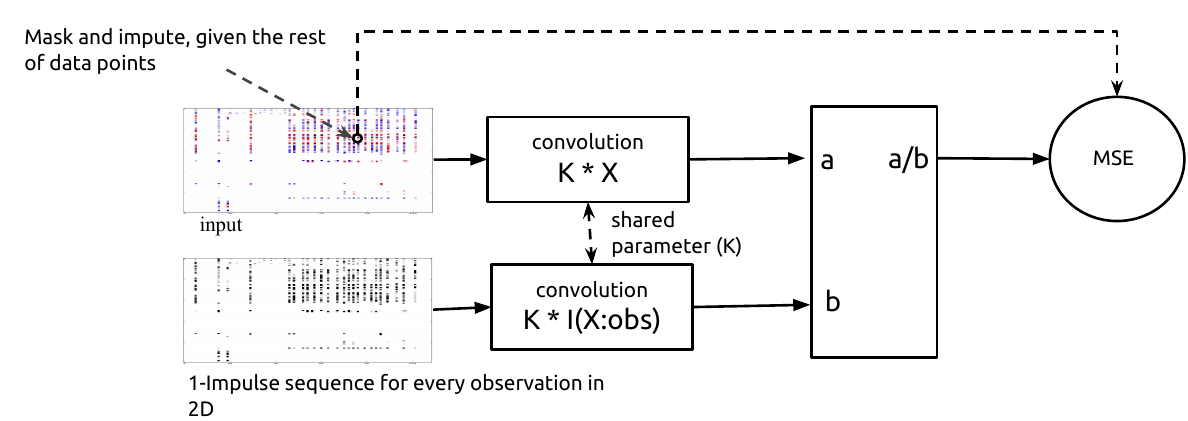}
\end{center}
\caption{Architecture for differentiable multivariate kernel regression.}\label{fig:deepkr}
\end{figure}

Figure \ref{fig:deepkr} shows the model for one output variable. Similar to the univariate training, for each time series, for each observation, we mask that observation and optimize the mean squared error of the true value compared to the predicted value using multivariate kernel. In our current formulation, we learn a separate $D \times (2M+1)$ sized kernel for each lab.

\begin{figure}
\begin{center}
\includegraphics[width=1\linewidth]{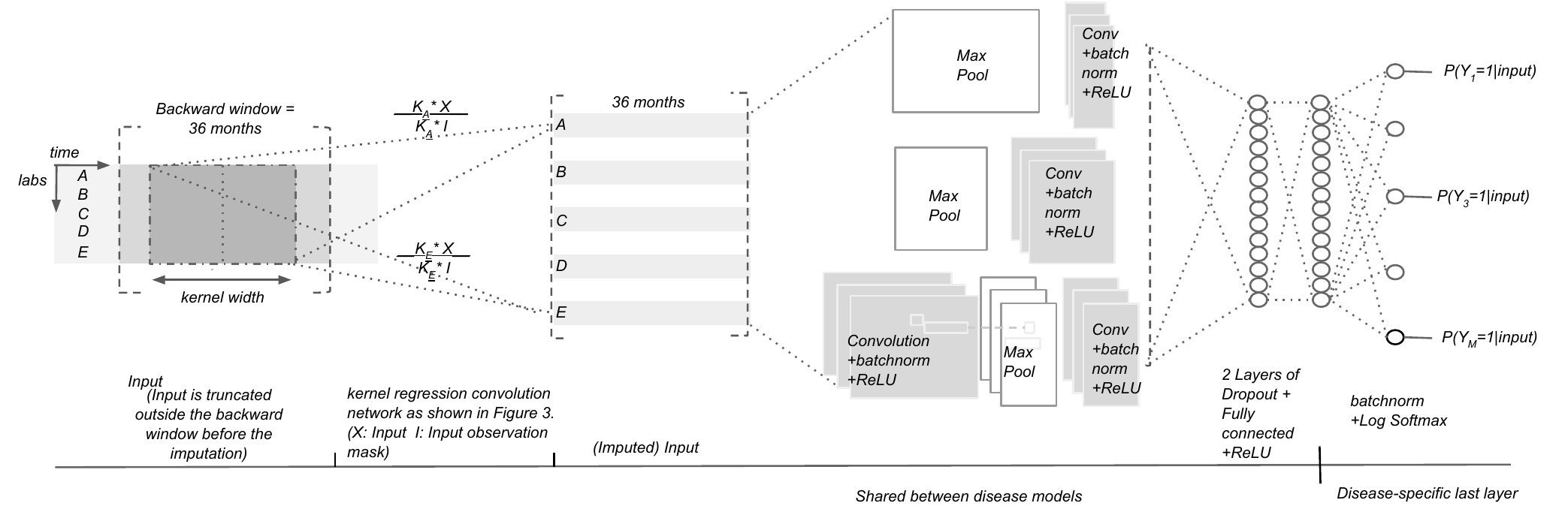}
\end{center}
\caption{Full network architecture for input to output prediction.}\label{fig:fullarch}
\end{figure}

Finally figure \ref{fig:fullarch} shows the full imputation and prediction architecture together. For each lab, the multivariate kernel is learned via pre-training and optimization of MSE for that lab. We then fix the imputation parameters, and train the consequent prediction network. We note that end-to-end training of the entire network (imputation and prediction) using the only prediction network's loss function will result in a different loss function than MSE for the imputation network. Joint training of the two networks, perhaps by optimizing both loss functions (negative log likelihood of predictions and mean squared error of imputations) is part of our future work. 

Finally, we note that at each time point, the input is truncated outside the backward window $before$ imputation, therefore no information from the future is affecting the prediction. 

% \subsubsection{Deep Model: Nonlinear Multivariate Kernel}
% In out next extension, we define the imputation function to be a general function of univariate kernel regressions.
% $$ x^d(t_{new})=f(\frac{k^1 \ast x^1}{k^1 \ast I^1}(t_{new}), \frac{k^2 \ast x^2}{k^2 \ast I^2}(t_{new})\dots ,\frac{k^D \ast x^D}{k^D \ast I^D})(t_{new}))$$
% % 
% This formulation allows us to share the univariate kernels $k^1$,.. $k^D$ between different outputs, and lowers the number of parameters from $D \times D \times (2M+1)$ to $D \times (2M+1)$ + $D \times P$ where $P$ is the number of parameters encoding the nonlinear multivariate interactions function $f$. 

% Figure \ref{fig:deepkr} on the right is an architecture for this model. This model allows us to handle the case where variables are measured anytime between $t-M$ to $t+M$. However, when 
% a variable is never observed in that period, univariate kernel regression is not defined. As we describe, the construction of $f$ should allow inference despite at the occurrence of missing values. 

\section{Experiments and Results}
\subsection{Data}
Our original dataset consisted of lab measurement and diagnosis information for 298,000 individuals. The lab measurements had the resolution of 1 month, and we used a backward window of 36 months for each prediction. We limited this paper's input to comprehensive lab panel plus cholesterol and bilirubin (together 18 lab types), which are currently recommended annually and covered by insurance companies. The name and code of labs used in our analysis is included in Table \ref{rmse_labs}. 
Each lab value was normalized by subtracting the mean and dividing by standard deviation across the entire dataset. 
We randomly divided individuals to a 100K training set, a 100K validation set, and a 98K test set. Validation set was used to select the best epoch/parameters for models and prediction results are presented on the test set, unseen during the training and validation.

Output corresponded to diagnosis information of these individuals. In our dataset, each disease diagnosis is recorded as an ICD9-CM (International Classification of Diseases, Ninth Revision, Clinical Modification) code. These codes are somewhat noisy, therefore we defined our prediction task carefully to improve the analysis quality as we describe next.

\subsection{Prediction Task Setup}\label{sec:res_pred}
Our goal is early diagnosis of diseases, for people who do not already have the disease. We required a 3 month gap between the end of the backward window (i.e. $t$), and the start of early diagnosis window. The purpose of the 3 month gap was to ensure that the clinical tests taken right before diagnosis of a disease would not allow our system to $cheat$ in the prediction of that disease. Each output label was defined as positive if the diagnosis code for the disease was observed in at least $2$ distinct months between $3$ to $3+24$ months after $t$. Using $24$ months helps alleviate the noisy label problem. Requiring at least $2$ observations of the noise also reduced the noise coming from "up-coding" physicians (physicians who report their wrong {\em suspected} diagnosis as a diagnosis). 
%A more fundamental solution to noisy labels via graphical models inference on the labels is in the list of our future work.
For each disease, we excluded individuals who already have the disease by time $t+3$. Our exclusion required only $1$ diagnosis record instead of $2$. This results in a more difficult, 
but also more clinically meaningful and interesting prediction task.

\subsection{Prediction Model Architecture Details}\label{sec:pred_spec}
The specific architectural choices for the shared part of the prediction network is as follows: We set the number of filters to be 8 for all convolution modules, with the kernel length 3(months) and step size of 1. Each Max-pooling module has the horizontal length of 3 and vertical length of 1, with step size of 3 in horizontal direction(i.e. no overlap). Each convolution module is followed by a batch normalization module \citep{ioffe2015batch} and then a ReLU nonlinearity \citep{nair2010rectified}. We have 2 fully connected layers (with 100 nodes each) after the concatenation of outputs of all convolution layers. Each of the fully connected layers are followed by a batch normalization layer and a ReLu nonlinearity layer. We also add one Dropout module \citep{srivastava2014dropout} (0.5 dropout probability) before each fully connected layer. After the last ReLu nonlinearity, corresponding to the output of the shared part of the network, for each 171 diseases we have the followings in order: A Dropout layer(0.5 dropout probability), a fully connected layer (of size 2 nodes corresponding to binary outcome), batch normalization layer and a Log Softmax Layer. Learning rate was selected from among the values $[0.001, 0.01, 0.05, 0.1, 1]$ using validation set average (over all diseases) Area Under ROC curve after 10 epochs. Value of $0.01$ was selected for the learning rate. Similarly learning rate decay of $0.95$ was selected for learning rate decay from the list $[0.8, 0.9, 0.95, 0.99]$. Training was done using stochastic gradient descent, over mini-batches of size 256. We implemented the architecture using Torch \citep{collobert2011torch7}.

\subsection{Data Augmentation}
During the training of the kernel regression imputation, we randomly perturbed each time series by adding Gaussian noise with standard deviation of 0.01 to each lab observation, and also randomly perturbed the time of each observation by a random jump drawn from a Gaussian distribution in either direction with standard deviation of 2 (we take the floor of the continuous value to determine the integer number of months to shift). We found this step to be especially important for learning robust imputation kernels.
% NOTE: said pre-training, rewrote to say training.

\subsection{Imputation Results}
First, we pre-trained the imputation layer by optimizing the mean squared error. For each observation, we masked the value, and asked the network (or our classic baselines of Gaussian Processes (\cite{rasmussen2006gaussian}) and classic Kernel regression) to predict the masked value given the rest of the observations. In case of the multivariate network we masked the data observed on the entire month during the training. 
We note that without data augmentation this method would not learn the value of the kernel at the origin (i.e. $t - t'=0$), and that's why randomly perturbing the $time$ of the observations by a small amount is essential.
Our baselines included univariate Gaussian Processes and univariate kernel regression. We used cross validation to select kernel family (Gaussian, Laplace and Triangular), kernel bandwidth, and in case of Gaussian Processes also the diagonal noise magnitude in the kernel matrix. 

Networks for each lab were trained independently. For this part of the analysis we used a random subset of 10,000 patients. 8,000 individuals were selected for training (or cross validation, in case of the baselines), and the other 2000 participated in evaluation.

\begin{figure}
\begin{center}
\includegraphics[width=0.35\linewidth]{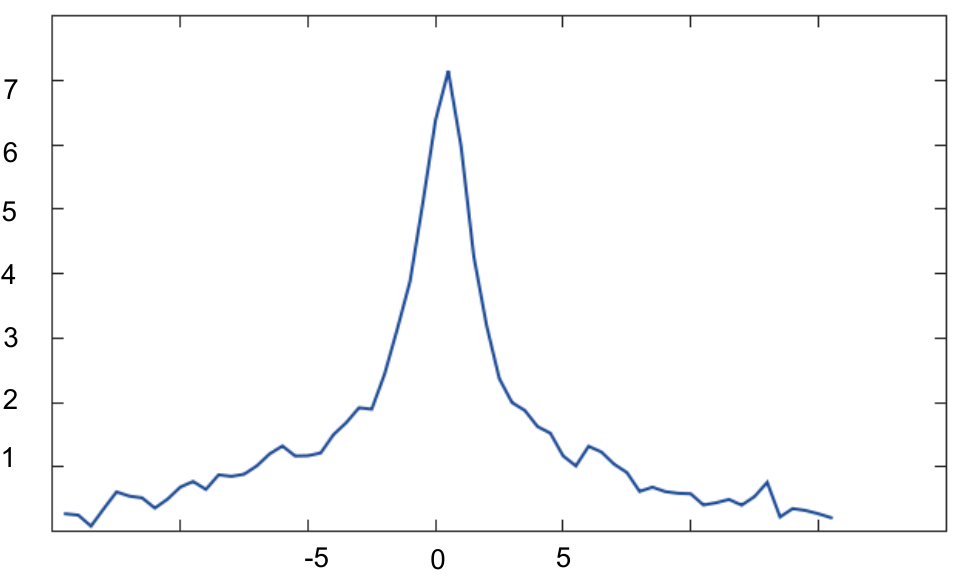}
\end{center}
\caption{The kernel learned for univariate kernel regression for Creatinine biomarker. The x axis indicates time with t =0 at the center, and the y axis is the magnitude of the kernel value around the origin.}\label{fig:univar_creat}
\end{figure}

\begin{figure}
\begin{center}
\includegraphics[width=1\linewidth]{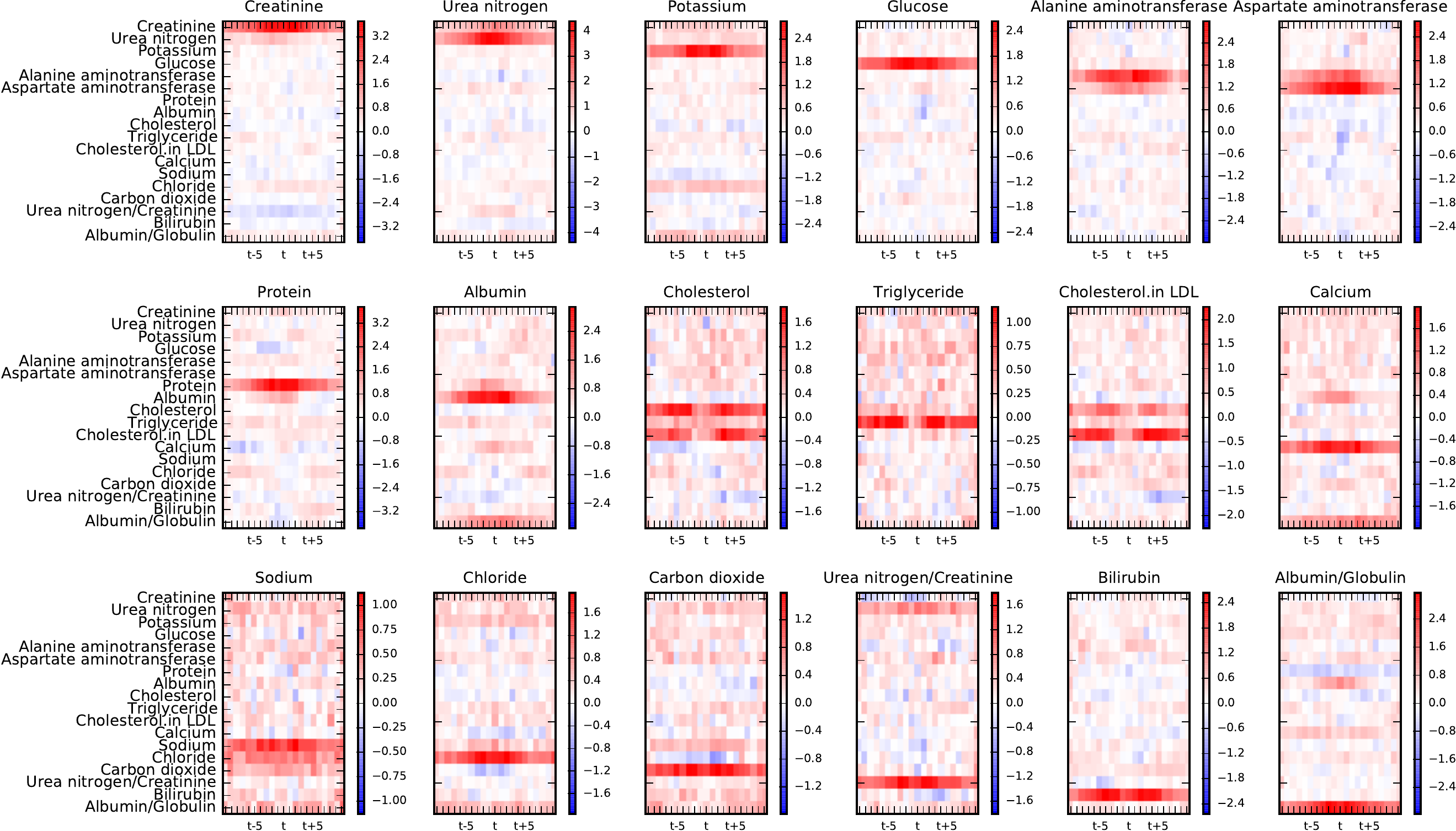}
\end{center}
\caption{The kernels learned for multivariate kernel regression network}\label{fig:multivar_kers}
\end{figure}

Table \ref{rmse_labs} shows the quality of the imputation on the lab values. Univariate models perform similarly, and since this is the result on the cohort which is already normalized, the univariate models are not reliable for many labs at all. But learning and using multivariate kernel model leads to a distinct improvement in the imputation quality. In Figure \ref{fig:univar_creat}, you can see the learned $univariate$ kernel for Creatinine lab. Trying to compose known families of kernel (i.e. Laplace or a mixture of Laplace kernels seems to fit well for the shape) to recover this form is not guaranteed to lead to the optimal data driven kernel. In Figure \ref{fig:multivar_kers} you can see the $multivariate$ kernels learned with our multivariate kernel regression framework. Our results indicate that the kernels capture the relationship between different variables well. Interesting to note is the lab value we purposefully did not discard, which is a ratio of two other lab values(Urea nitrogen/Creatinine). Our formulation of multivariate kernel regression only allows linear construction at this level of depth, and we see that the ratio is approximated with positive weight for numerator(Urea nitrogen) and negative weight for denominator(Creatinine). 

\begin{table}
\caption{RMSE for different lab values and different models. Models are: GP(Univariate Gaussian Processes), KR(Kernel Regression), ConvKR(Univariate convolution formulation of KR), and ConvKR multivariate.}
\label{rmse_labs}
\begin{center}
\begin{tabular}{lllll}
\multicolumn{1}{c}{\bf Lab}&\multicolumn{1}{c}{\bf GP}   &\multicolumn{1}{c}{\bf KR univar} & \multicolumn{1}{c}{\bf Conv KR univar}  &\multicolumn{1}{c}{\bf Conv KR multivar}
\\ \hline \\
Creatinine	&0.397	&	0.406	&	0.433	&	0.096	\\
Urea nitrogen	&0.449	&	0.457	&	0.465	&	0.131	\\
Potassium	&0.995	&	1.011	&	1.010	&	0.170	\\
Glucose	&0.716	&	0.709	&	0.690	&	0.118	\\
Alanine aminotransferase	&	0.653	&	0.677	&	0.679	&	0.127	\\
Aspartate aminotransferase	&	0.708	&	0.720	&	0.710	&	0.130	\\
Protein	&1.142	&	1.194	&	1.220	&	0.206	\\
Albumin	&1.092	&	1.128	&	1.120	&	0.263	\\
Cholesterol	&0.621	&	0.631	&	0.651	&	0.118	\\
Triglyceride	&0.640	&	0.633	&	0.696	&	0.104	\\
Cholesterol.in LDL	&	0.640	&	0.649	&	0.648	&	0.108	\\
Calcium	&1.614	&	1.652	&	1.703	&	0.260	\\
Sodium	&0.722	&	0.717	&	0.742	&	0.139	\\
Chloride	&0.672	&	0.674	&	0.688	&	0.113	\\
Carbon dioxide	&0.782	&	0.783	&	0.782	&	0.131	\\
Urea nitrogen/Creatinine	&	0.601	&	0.606	&	0.600	&	0.075	\\
Bilirubin	&	0.667	&	0.687	&	0.678	&	0.105	\\
Albumin/Globulin	&0.586	&	0.601	&	0.636	&	0.112	\\
\end{tabular}
\end{center}
\end{table}

%\begin{figure}[h]
%\begin{center}
%\includegraphics[width=1\linewidth]{img/rmses-crop.pdf}
%\end{center}
%\caption{Comparing RMSE for different lab values and different models. Each lab titles are accompanies with the LOINC code, corresponding to the exact chemical ID. Models are: GP(Univariate Gaussian Processes), KR(Kernel Regression), ConvKR(Univariate convolution formulation of KR), and ConvKR multivariate.}\label{fig:rmse_labs}
%\end{figure}

% \begin{table}[t]
% \caption{Sample table title}
% \label{sample-table}
% \begin{center}
% \begin{tabular}{ll}
% \multicolumn{1}{c}{\bf PART}  &\multicolumn{1}{c}{\bf DESCRIPTION}
% \\ \hline \\
% Dendrite         &Input terminal \\
% Axon             &Output terminal \\
% Soma             &Cell body (contains cell nucleus) \\
% \end{tabular}
% \end{center}
% \end{table}

Kernel based imputation method has a property where the kernel is symmetric, looking into past and future for the imputation. Our training method optimizes the prediction of observations within and at the border of each lab time series, therefore allowing the kernels to adjust to the border cases where only past(or future) data is available. However, in the consequent disease prediction model we only use a 3-year backward window which is shorter than the typical time-span on which imputation was trained. While we do not explore in this paper, the interplay between the spans of the kernels, the quality of imputation at border cases, and their effect on sub-sequent disease prediction task is an interesting direction for further analysis.

\subsection{Prediction Results} 
Figure \ref{fig:aucs}, and table \ref{tab:auc_table} in the Supplementary section show the area under ROC curve results of predicting each disease on the test set. As baselines we compare the results to multilayer perceptron (MLP) over the entire observations within the 36 month backward window, and logistic regression on maximum value of all 18 lab values over 36 months backward window. For each model, we compared three imputation setting: Raw input, without imputing unobserved values; Imputed input only; and a 2-channel input, composed of imputed input, next to binary observation mask. Results shown in Figure \ref{fig:aucs} are the best AUCs achieved on any of the imputation settings, per model. 

Our multilayer perceptron baseline had 2 hidden layers shared across all 171 diseases (100 hidden nodes each). Each disease is predicted using a logistic
function of the last hidden layer, with its own parameters (implemented as Log Softmax in Torch environment). Batch normalization \citep{ioffe2015batch} was used after every hidden layer, and Dropout (with probability 0.5) was used before each hidden layer. We used cross-validation to optimize learning rate and learning rate decay for the baseline models. We also selected optimum learning rate and learning rate decay parameters for the convolution network via cross-validation, but fixed the architecture parameters to those described in section 2. As in the case with convolution network, we used weighted negative log-likelihood loss function to train the baselines.

In figure \ref{fig:aucs_full} and table \ref{tab:auc_table_full} in Supplementary section, we show the disease classification AUC results on the full set of experiments, comparing different imputation and prediction methods. By imputation, we discard health-care utilization, which is a predictive signal. So it is expected that just using imputation would lower the early detection accuracy, and we observe this in our results as well. However, when we use two separate channels (imputed signal and binary observation mask), the results are comparable to the prediction on the unimputed input, which indicates that the our imputation layer is correctly separating biological signals from the utilization. The trends learned on the imputed channel are more interesting to medical research field, which studies core biological processes, while clinical intervention field is interested in any model that gives better predictive result, using all the signals including the utilization patterns.

Clinically interesting to note is how well can different diseases be diagnosed at least 3 month in advance, and how many diseases can be detected with much better accuracy compared to current practices, by using the convolution method. In particular, heart failure, severe kidney diseases and liver problems, diabetes and hormone related conditions and prostate cancer are among the diseases which are well detectable early, from only 18 common lab measurements tracked in the past 3 years. Additionally, for patients with multiple existing diseases, side-effects of different medications in addition to their conditions can trigger other unexpected diseases. Monitoring the risks of all diseases is often neglected in the clinics today and our model is a reasonable solution for this task. 

\begin{figure}
\begin{center}
\includegraphics[width=1\linewidth]{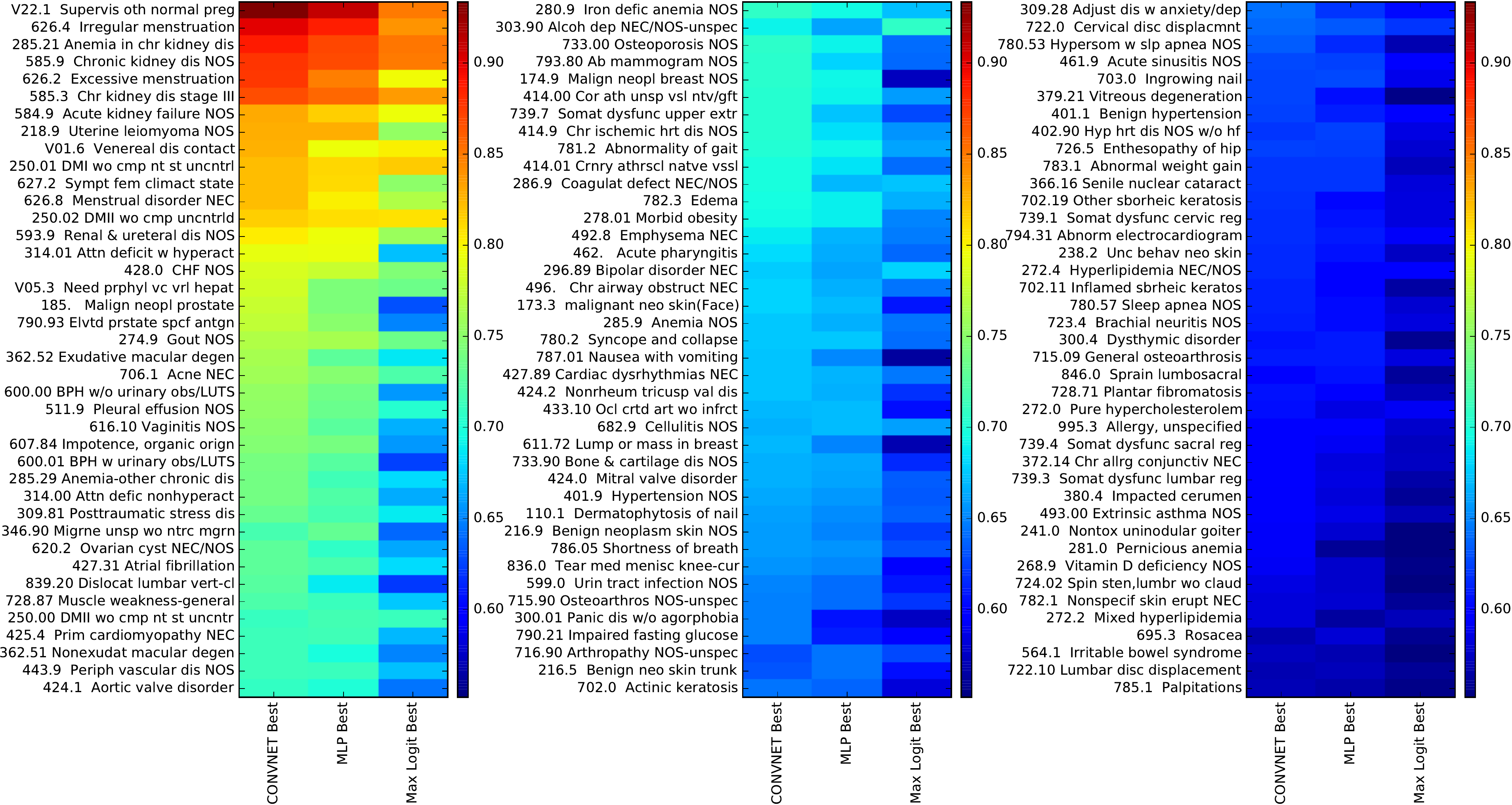}% aucs1-crop.pdf}
\end{center}
\caption{Area Under ROC curve for each disease, comparing the held out test score on our Convolution vs. Multi-layered Perceptron, vs. Logistic Regression over the maximum observed lab in the past 3 years. The actual AUC values are included in table \ref{tab:auc_table} }\label{fig:aucs}
\end{figure}
	
\section{Conclusion}
In this work, we presented the first large scale application of convolutional neural networks for discovery of early temporal disease signatures for the task of disease prediction. We presented a novel approach to nonparametric imputation, which is essential to learning disease signatures that are biologically valid. Our results show significant improvement in the quality of early diagnosis, compared to methods currently used in most of the medical and clinical world, only using 18 lab measurement over the past 3 years. 

Our results indicate that onset of many diseases, including major heart, kidney, and liver diseases, prostate cancer, and diabetes are predictable with high quality in advance. For many of these diseases, early detection even by a few months can lead to significant gains in effectiveness of treatment, quality of life of the patients and their families, and reduction of financial burden on the healthcare systems. Our method also enables large-scale intervention programs to target the most high-risk population better than available baselines, therefore increasing the cost-effectiveness and applicability of the programs. Finally, for every disease presented, detailed analysis of the discovered predictive temporal patterns can lead to new medical insights, and is part of our future work.

% Additionally, for patients with multiple existing diseases who suffer from side-effects of different drugs, monitoring the risks of other diseases is often neglected. 
%Further analysis of the recovered multiple resolution patterns on the lab measurements per disease is part of our future work.

\section*{Acknowledgements}
The authors gratefully acknowledge support by Independence Blue Cross. The Tesla K40s used for this research were donated by the NVIDIA Corporation.

\bibliography{paper}
\bibliographystyle{iclr2016_conference}
\beginsupplement

\section{Supplementary Figure}
\begin{figure}
\begin{center}
\includegraphics[width=1\linewidth]{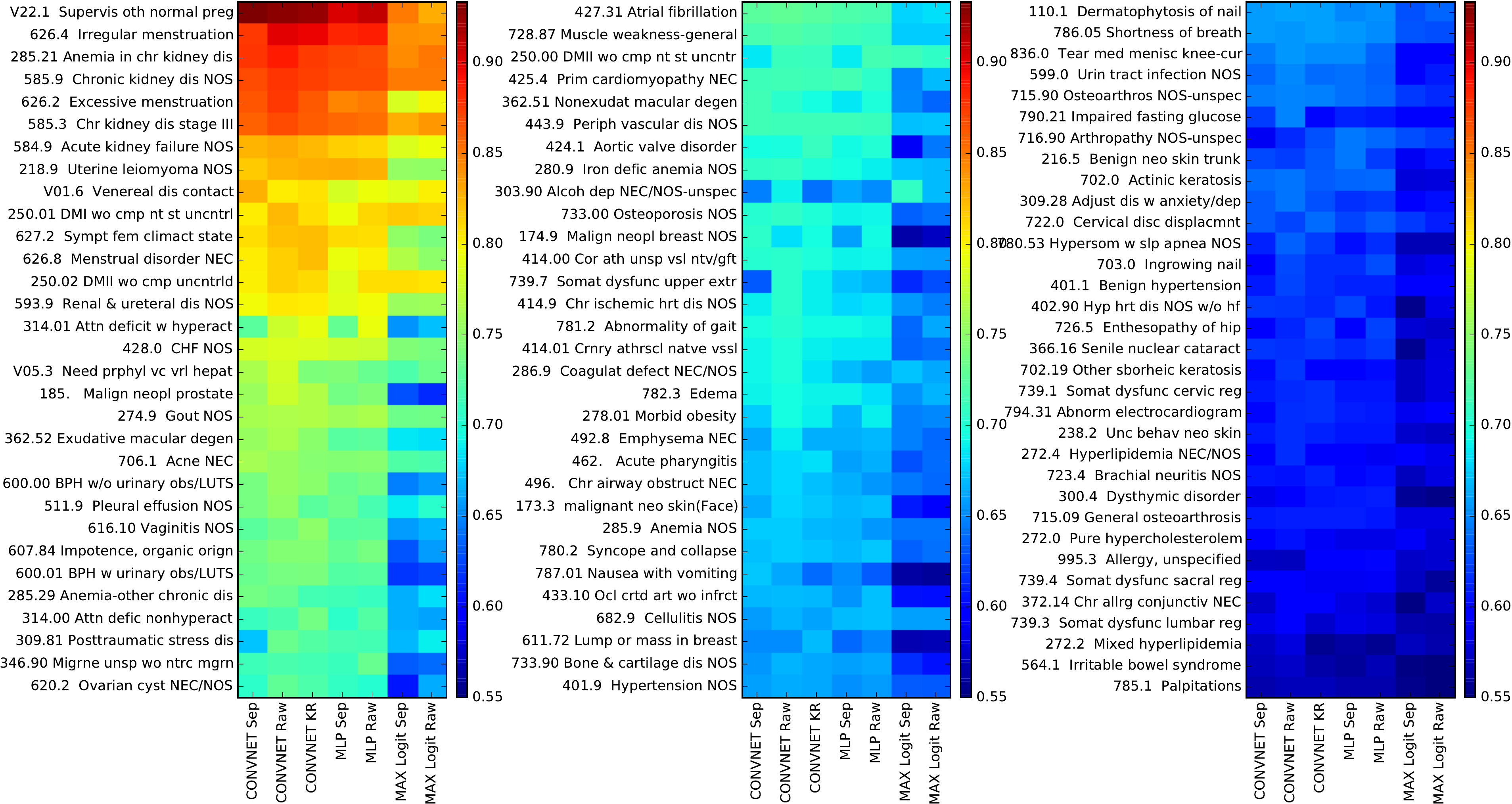}
\end{center}
\caption{Area Under ROC curve for each disease, comparing the held out test score on our Convolution vs. Multi-layered Perceptron, vs. Logistic Regression over the maximum observed lab in the past 3 years, and Comparing different imputation methods (Sep = input is two channel, composed of imputed input + observation mask), (Raw = input not imputed, and processed as raw leaving unobserved months as zero), and (KR = imputed input only). The actual AUC values are included in table \ref{tab:auc_table_full}}\label{fig:aucs_full}
\end{figure}

\begin{center}
\begin{longtable}{llll}
\caption{Area Under ROC curve for each disease, comparing the held out test score on our Convolution(Convnet) vs. Multi-layered Perceptron(MLP), vs. Logistic Regression(Logit) over the maximum observed lab in the past 3 years, corresponding to Figure \ref{fig:aucs}}\label{tab:auc_table}\\

\multicolumn{1}{c}{\bf Disease ICD9 Code and Label} & \multicolumn{1}{c}{\bf Convnet AUC}   & \multicolumn{1}{c}{\bf MLP AUC} & \multicolumn{1}{c}{\bf Logit AUC}
\\\hline \\
\endfirsthead

\multicolumn{4}{r}{{Continued on next page}} \\
\endfoot
\multicolumn{4}{r}{} \\
\endlastfoot
V22.1  Supervis oth normal preg  & 0.933 & 0.911 & 0.849 \\
626.4  Irregular menstruation  & 0.899 & 0.886 & 0.839 \\
285.21 Anemia in chr kidney dis  & 0.888 & 0.870 & 0.850 \\
585.9  Chronic kidney dis NOS  & 0.876 & 0.868 & 0.849 \\
626.2  Excessive menstruation  & 0.875 & 0.848 & 0.797 \\
585.3  Chr kidney dis stage III  & 0.867 & 0.856 & 0.836 \\
584.9  Acute kidney failure NOS  & 0.829 & 0.815 & 0.795 \\
218.9  Uterine leiomyoma NOS  & 0.828 & 0.828 & 0.753 \\
V01.6  Venereal dis contact  & 0.826 & 0.795 & 0.801 \\
250.01 DMI wo cmp nt st uncntrl  & 0.822 & 0.811 & 0.816 \\
627.2  Sympt fem climact state  & 0.821 & 0.809 & 0.751 \\
626.8  Menstrual disorder NEC  & 0.821 & 0.801 & 0.767 \\
250.02 DMII wo cmp uncntrld  & 0.814 & 0.808 & 0.807 \\
593.9  Renal and ureteral dis NOS  & 0.803 & 0.796 & 0.756 \\
314.01 Attn deficit w hyperact  & 0.792 & 0.793 & 0.671 \\
428.0  CHF NOS  & 0.785 & 0.778 & 0.744 \\
V05.3  Need prphyl vc vrl hepat  & 0.782 & 0.744 & 0.735 \\
185.   Malign neopl prostate  & 0.778 & 0.743 & 0.629 \\
790.93 Elvtd prstate spcf antgn  & 0.775 & 0.749 & 0.648 \\
274.9  Gout NOS  & 0.767 & 0.763 & 0.736 \\
362.52 Exudative macular degen  & 0.763 & 0.728 & 0.686 \\
706.1  Acne NEC  & 0.760 & 0.747 & 0.721 \\
600.00 BPH w/o urinary obs/LUTS  & 0.754 & 0.737 & 0.657 \\
511.9  Pleural effusion NOS  & 0.751 & 0.734 & 0.703 \\
616.10 Vaginitis NOS  & 0.749 & 0.726 & 0.665 \\
607.84 Impotence, organic orign  & 0.746 & 0.740 & 0.657 \\
600.01 BPH w urinary obs/LUTS  & 0.740 & 0.724 & 0.623 \\
285.29 Anemia-other chronic dis  & 0.738 & 0.714 & 0.682 \\
314.00 Attn defic nonhyperact  & 0.738 & 0.722 & 0.664 \\
309.81 Posttraumatic stress dis  & 0.733 & 0.717 & 0.688 \\
346.90 Migrne unsp wo ntrc mgrn  & 0.718 & 0.732 & 0.639 \\
620.2  Ovarian cyst NEC/NOS  & 0.728 & 0.707 & 0.663 \\
427.31 Atrial fibrillation  & 0.728 & 0.720 & 0.682 \\
839.20 Dislocat lumbar vert-cl  & 0.725 & 0.688 & 0.621 \\
728.87 Muscle weakness-general  & 0.721 & 0.712 & 0.675 \\
250.00 DMII wo cmp nt st uncntr  & 0.710 & 0.716 & 0.712 \\
425.4  Prim cardiomyopathy NEC  & 0.713 & 0.715 & 0.668 \\
362.51 Nonexudat macular degen  & 0.713 & 0.697 & 0.649 \\
443.9  Periph vascular dis NOS  & 0.713 & 0.712 & 0.672 \\
424.1  Aortic valve disorder  & 0.709 & 0.700 & 0.643 \\
280.9  Iron defic anemia NOS  & 0.708 & 0.694 & 0.671 \\
303.90 Alcoh dep NEC/NOS-unspec  & 0.690 & 0.661 & 0.707 \\
733.00 Osteoporosis NOS  & 0.706 & 0.691 & 0.641 \\
793.80 Ab mammogram NOS  & 0.704 & 0.679 & 0.639 \\
174.9  Malign neopl breast NOS  & 0.704 & 0.692 & 0.573 \\
414.00 Cor ath unsp vsl ntv/gft  & 0.703 & 0.691 & 0.659 \\
739.7  Somat dysfunc upper extr  & 0.703 & 0.672 & 0.627 \\
414.9  Chr ischemic hrt dis NOS  & 0.703 & 0.683 & 0.655 \\
781.2  Abnormality of gait  & 0.702 & 0.692 & 0.661 \\
414.01 Crnry athrscl natve vssl  & 0.697 & 0.685 & 0.640 \\
286.9  Coagulat defect NEC/NOS  & 0.697 & 0.669 & 0.672 \\
782.3  Edema  & 0.695 & 0.689 & 0.666 \\
278.01 Morbid obesity  & 0.694 & 0.689 & 0.649 \\
492.8  Emphysema NEC  & 0.687 & 0.667 & 0.646 \\
462.   Acute pharyngitis  & 0.682 & 0.664 & 0.639 \\
296.89 Bipolar disorder NEC  & 0.675 & 0.660 & 0.680 \\
496.   Chr airway obstruct NEC  & 0.679 & 0.665 & 0.643 \\
173.3  malignant neo skin(Face)    & 0.678 & 0.670 & 0.608 \\
285.9  Anemia NOS  & 0.675 & 0.665 & 0.642 \\
780.2  Syncope and collapse  & 0.675 & 0.674 & 0.640 \\
787.01 Nausea with vomiting  & 0.673 & 0.651 & 0.563 \\
427.89 Cardiac dysrhythmias NEC  & 0.673 & 0.667 & 0.645 \\
424.2  Nonrheum tricusp val dis  & 0.672 & 0.666 & 0.617 \\
433.10 Ocl crtd art wo infrct  & 0.668 & 0.670 & 0.605 \\
682.9  Cellulitis NOS  & 0.665 & 0.669 & 0.659 \\
611.72 Lump or mass in breast  & 0.668 & 0.649 & 0.568 \\
733.90 Bone and cartilage dis NOS  & 0.666 & 0.662 & 0.615 \\
424.0  Mitral valve disorder  & 0.666 & 0.661 & 0.634 \\
401.9  Hypertension NOS  & 0.663 & 0.656 & 0.632 \\
110.1  Dermatophytosis of nail  & 0.659 & 0.652 & 0.636 \\
216.9  Benign neoplasm skin NOS  & 0.659 & 0.649 & 0.623 \\
786.05 Shortness of breath  & 0.658 & 0.655 & 0.630 \\
836.0  Tear med menisc knee-cur  & 0.656 & 0.651 & 0.599 \\
599.0  Urin tract infection NOS  & 0.650 & 0.641 & 0.608 \\
715.90 Osteoarthros NOS-unspec  & 0.648 & 0.641 & 0.619 \\
300.01 Panic dis w/o agorphobia  & 0.647 & 0.605 & 0.575 \\
790.21 Impaired fasting glucose  & 0.647 & 0.610 & 0.596 \\
716.90 Arthropathy NOS-unspec  & 0.627 & 0.643 & 0.626 \\
216.5  Benign neo skin trunk  & 0.631 & 0.643 & 0.604 \\
702.0  Actinic keratosis  & 0.642 & 0.637 & 0.582 \\
309.28 Adjust dis w anxiety/dep  & 0.642 & 0.619 & 0.602 \\
722.0  Cervical disc displacmnt  & 0.638 & 0.633 & 0.620 \\
780.53 Hypersom w slp apnea NOS  & 0.634 & 0.614 & 0.569 \\
461.9  Acute sinusitis NOS  & 0.626 & 0.623 & 0.597 \\
703.0  Ingrowing nail  & 0.625 & 0.626 & 0.588 \\
379.21 Vitreous degeneration  & 0.625 & 0.605 & 0.555 \\
401.1  Benign hypertension  & 0.624 & 0.611 & 0.597 \\
402.90 Hyp hrt dis NOS w/o hf  & 0.619 & 0.624 & 0.585 \\
726.5  Enthesopathy of hip  & 0.623 & 0.622 & 0.579 \\
783.1  Abnormal weight gain  & 0.619 & 0.620 & 0.572 \\
366.16 Senile nuclear cataract  & 0.619 & 0.619 & 0.582 \\
702.19 Other sborheic keratosis  & 0.618 & 0.602 & 0.583 \\
739.1  Somat dysfunc cervic reg  & 0.617 & 0.606 & 0.584 \\
794.31 Abnorm electrocardiogram  & 0.616 & 0.608 & 0.592 \\
238.2  Unc behav neo skin  & 0.615 & 0.606 & 0.575 \\
272.4  Hyperlipidemia NEC/NOS  & 0.614 & 0.595 & 0.597 \\
702.11 Inflamed sbrheic keratos  & 0.612 & 0.595 & 0.565 \\
780.57 Sleep apnea NOS  & 0.611 & 0.603 & 0.578 \\
723.4  Brachial neuritis NOS  & 0.610 & 0.603 & 0.584 \\
300.4  Dysthymic disorder  & 0.606 & 0.609 & 0.557 \\
715.09 General osteoarthrosis  & 0.608 & 0.608 & 0.584 \\
846.0  Sprain lumbosacral  & 0.596 & 0.606 & 0.561 \\
728.71 Plantar fibromatosis  & 0.605 & 0.599 & 0.569 \\
272.0  Pure hypercholesterolem  & 0.605 & 0.585 & 0.590 \\
995.3  Allergy, unspecified  & 0.596 & 0.600 & 0.577 \\
739.4  Somat dysfunc sacral reg  & 0.597 & 0.591 & 0.572 \\
372.14 Chr allrg conjunctiv NEC  & 0.596 & 0.584 & 0.574 \\
739.3  Somat dysfunc lumbar reg  & 0.595 & 0.585 & 0.565 \\
380.4  Impacted cerumen  & 0.594 & 0.582 & 0.560 \\
493.00 Extrinsic asthma NOS  & 0.594 & 0.586 & 0.570 \\
241.0  Nontox uninodular goiter  & 0.593 & 0.579 & 0.552 \\
281.0  Pernicious anemia  & 0.592 & 0.560 & 0.552 \\
268.9  Vitamin D deficiency NOS  & 0.591 & 0.578 & 0.555 \\
724.02 Spin sten,lumbr wo claud  & 0.585 & 0.578 & 0.552 \\
782.1  Nonspecif skin erupt NEC  & 0.582 & 0.580 & 0.559 \\
272.2  Mixed hyperlipidemia  & 0.582 & 0.562 & 0.571 \\
695.3  Rosacea  & 0.566 & 0.581 & 0.558 \\
564.1  Irritable bowel syndrome  & 0.574 & 0.569 & 0.551 \\
722.10 Lumbar disc displacement  & 0.569 & 0.572 & 0.559 \\
785.1  Palpitations  & 0.570 & 0.565 & 0.555 \\
\hline
\end{longtable}
\end{center}

\begin{center}
\begin{longtable}{llllllll}
\caption{Area Under ROC curve for each disease, comparing the held out test score on our Convolution (C) vs. Multi-layered Perceptron(M), vs. Logistic Regression over the maximum observed lab in the past 3 years(L), and Comparing different imputation methods (SP = input is two channel, composed of imputed input + observation mask), (RW = input not imputed, and processed as raw leaving unobserved months as zero), and (KR = imputed input only) This table exactly corresponds to Figure \ref{fig:aucs_full}}\label{tab:auc_table_full}\\

\multicolumn{1}{c}{\bf Disease ICD9 Code and Label} & \multicolumn{1}{c}{\bf C-SP}   & \multicolumn{1}{c}{\bf C-RW} & \multicolumn{1}{c}{\bf C-KR} & \multicolumn{1}{c}{\bf M-SP} & \multicolumn{1}{c}{\bf M-RW} & \multicolumn{1}{c}{\bf L-SP} & \multicolumn{1}{c}{\bf L-RW} 
\\\hline \\
\endfirsthead

\multicolumn{8}{r}{{Continued on next page}} \\
\endfoot
\multicolumn{8}{r}{} \\
\endlastfoot
V22.1  Supervis oth normal preg  & 0.933 & 0.928 & 0.924 & 0.902 & 0.911 & 0.849 & 0.828 \\
626.4  Irregular menstruation  & 0.876 & 0.899 & 0.896 & 0.882 & 0.886 & 0.839 & 0.838 \\
285.21 Anemia in chr kidney dis  & 0.877 & 0.888 & 0.873 & 0.870 & 0.866 & 0.841 & 0.850 \\
585.9  Chronic kidney dis NOS  & 0.868 & 0.876 & 0.872 & 0.868 & 0.868 & 0.848 & 0.849 \\
626.2  Excessive menstruation  & 0.863 & 0.875 & 0.862 & 0.844 & 0.848 & 0.784 & 0.797 \\
585.3  Chr kidney dis stage III  & 0.859 & 0.867 & 0.860 & 0.856 & 0.854 & 0.828 & 0.836 \\
584.9  Acute kidney failure NOS  & 0.825 & 0.829 & 0.823 & 0.815 & 0.812 & 0.785 & 0.795 \\
218.9  Uterine leiomyoma NOS  & 0.815 & 0.825 & 0.828 & 0.828 & 0.826 & 0.753 & 0.751 \\
V01.6  Venereal dis contact  & 0.826 & 0.802 & 0.806 & 0.782 & 0.795 & 0.788 & 0.801 \\
250.01 DMI wo cmp nt st uncntrl  & 0.801 & 0.822 & 0.808 & 0.795 & 0.811 & 0.816 & 0.812 \\
627.2  Sympt fem climact state  & 0.808 & 0.820 & 0.821 & 0.809 & 0.808 & 0.751 & 0.740 \\
626.8  Menstrual disorder NEC  & 0.801 & 0.815 & 0.821 & 0.794 & 0.801 & 0.767 & 0.749 \\
250.02 DMII wo cmp uncntrld  & 0.801 & 0.814 & 0.810 & 0.786 & 0.808 & 0.807 & 0.805 \\
593.9  Renal and ureteral dis NOS  & 0.796 & 0.803 & 0.801 & 0.788 & 0.796 & 0.755 & 0.756 \\
314.01 Attn deficit w hyperact  & 0.726 & 0.778 & 0.792 & 0.731 & 0.793 & 0.654 & 0.671 \\
428.0  CHF NOS  & 0.783 & 0.785 & 0.783 & 0.778 & 0.777 & 0.744 & 0.739 \\
V05.3  Need prphyl vc vrl hepat  & 0.763 & 0.782 & 0.741 & 0.744 & 0.731 & 0.719 & 0.735 \\
185.   Malign neopl prostate  & 0.755 & 0.778 & 0.766 & 0.737 & 0.743 & 0.629 & 0.613 \\
274.9  Gout NOS  & 0.761 & 0.764 & 0.767 & 0.758 & 0.763 & 0.735 & 0.736 \\
362.52 Exudative macular degen  & 0.754 & 0.763 & 0.748 & 0.723 & 0.728 & 0.686 & 0.682 \\
706.1  Acne NEC  & 0.760 & 0.755 & 0.746 & 0.745 & 0.747 & 0.721 & 0.718 \\
600.00 BPH w/o urinary obs/LUTS  & 0.740 & 0.754 & 0.748 & 0.731 & 0.737 & 0.646 & 0.657 \\
511.9  Pleural effusion NOS  & 0.739 & 0.751 & 0.726 & 0.734 & 0.716 & 0.687 & 0.703 \\
616.10 Vaginitis NOS  & 0.725 & 0.736 & 0.749 & 0.725 & 0.726 & 0.656 & 0.665 \\
607.84 Impotence, organic orign  & 0.735 & 0.745 & 0.746 & 0.730 & 0.740 & 0.629 & 0.657 \\
600.01 BPH w urinary obs/LUTS  & 0.732 & 0.739 & 0.740 & 0.723 & 0.724 & 0.618 & 0.623 \\
285.29 Anemia-other chronic dis  & 0.738 & 0.732 & 0.715 & 0.714 & 0.708 & 0.664 & 0.682 \\
314.00 Attn defic nonhyperact  & 0.709 & 0.718 & 0.738 & 0.704 & 0.722 & 0.664 & 0.660 \\
309.81 Posttraumatic stress dis  & 0.672 & 0.733 & 0.722 & 0.717 & 0.714 & 0.667 & 0.688 \\
346.90 Migrne unsp wo ntrc mgrn  & 0.712 & 0.718 & 0.715 & 0.710 & 0.732 & 0.632 & 0.639 \\
620.2  Ovarian cyst NEC/NOS  & 0.705 & 0.728 & 0.719 & 0.707 & 0.702 & 0.606 & 0.663 \\
427.31 Atrial fibrillation  & 0.725 & 0.728 & 0.724 & 0.720 & 0.716 & 0.678 & 0.682 \\
728.87 Muscle weakness-general  & 0.716 & 0.721 & 0.716 & 0.712 & 0.709 & 0.674 & 0.675 \\
250.00 DMII wo cmp nt st uncntr  & 0.685 & 0.710 & 0.709 & 0.687 & 0.716 & 0.712 & 0.706 \\
425.4  Prim cardiomyopathy NEC  & 0.712 & 0.713 & 0.712 & 0.715 & 0.705 & 0.648 & 0.668 \\
362.51 Nonexudat macular degen  & 0.713 & 0.703 & 0.694 & 0.685 & 0.697 & 0.649 & 0.635 \\
443.9  Periph vascular dis NOS  & 0.713 & 0.712 & 0.712 & 0.712 & 0.711 & 0.670 & 0.672 \\
424.1  Aortic valve disorder  & 0.694 & 0.696 & 0.709 & 0.692 & 0.700 & 0.590 & 0.643 \\
280.9  Iron defic anemia NOS  & 0.707 & 0.708 & 0.698 & 0.694 & 0.688 & 0.671 & 0.668 \\
303.90 Alcoh dep NEC/NOS-unspec  & 0.646 & 0.690 & 0.641 & 0.661 & 0.651 & 0.707 & 0.669 \\
733.00 Osteoporosis NOS  & 0.702 & 0.706 & 0.698 & 0.690 & 0.691 & 0.634 & 0.641 \\
174.9  Malign neopl breast NOS  & 0.704 & 0.683 & 0.695 & 0.658 & 0.692 & 0.563 & 0.573 \\
414.00 Cor ath unsp vsl ntv/gft  & 0.701 & 0.703 & 0.697 & 0.689 & 0.691 & 0.659 & 0.656 \\
739.7  Somat dysfunc upper extr  & 0.631 & 0.703 & 0.688 & 0.672 & 0.666 & 0.613 & 0.627 \\
414.9  Chr ischemic hrt dis NOS  & 0.687 & 0.703 & 0.689 & 0.679 & 0.683 & 0.655 & 0.645 \\
781.2  Abnormality of gait  & 0.696 & 0.702 & 0.693 & 0.692 & 0.690 & 0.638 & 0.661 \\
414.01 Crnry athrscl natve vssl  & 0.691 & 0.697 & 0.688 & 0.685 & 0.685 & 0.636 & 0.640 \\
286.9  Coagulat defect NEC/NOS  & 0.691 & 0.697 & 0.684 & 0.669 & 0.658 & 0.672 & 0.662 \\
782.3  Edema  & 0.690 & 0.695 & 0.689 & 0.689 & 0.682 & 0.654 & 0.666 \\
278.01 Morbid obesity  & 0.674 & 0.694 & 0.688 & 0.665 & 0.689 & 0.646 & 0.649 \\
492.8  Emphysema NEC  & 0.661 & 0.687 & 0.664 & 0.663 & 0.667 & 0.646 & 0.637 \\
462.   Acute pharyngitis  & 0.671 & 0.679 & 0.682 & 0.658 & 0.664 & 0.628 & 0.639 \\
496.   Chr airway obstruct NEC  & 0.670 & 0.679 & 0.671 & 0.665 & 0.655 & 0.643 & 0.637 \\
173.3  malignant neo skin(Face)    & 0.663 & 0.678 & 0.674 & 0.668 & 0.670 & 0.608 & 0.592 \\
285.9  Anemia NOS  & 0.675 & 0.674 & 0.666 & 0.665 & 0.655 & 0.641 & 0.642 \\
780.2  Syncope and collapse  & 0.669 & 0.675 & 0.671 & 0.667 & 0.674 & 0.634 & 0.640 \\
787.01 Nausea with vomiting  & 0.673 & 0.661 & 0.639 & 0.651 & 0.633 & 0.563 & 0.560 \\
433.10 Ocl crtd art wo infrct  & 0.667 & 0.668 & 0.668 & 0.654 & 0.670 & 0.605 & 0.603 \\
682.9  Cellulitis NOS  & 0.656 & 0.665 & 0.660 & 0.669 & 0.669 & 0.658 & 0.659 \\
611.72 Lump or mass in breast  & 0.651 & 0.651 & 0.668 & 0.637 & 0.649 & 0.566 & 0.568 \\
733.90 Bone and cartilage dis NOS  & 0.656 & 0.666 & 0.661 & 0.658 & 0.662 & 0.615 & 0.604 \\
401.9  Hypertension NOS  & 0.659 & 0.663 & 0.661 & 0.652 & 0.656 & 0.632 & 0.631 \\
110.1  Dermatophytosis of nail  & 0.656 & 0.659 & 0.658 & 0.650 & 0.652 & 0.628 & 0.636 \\
786.05 Shortness of breath  & 0.657 & 0.656 & 0.658 & 0.655 & 0.650 & 0.630 & 0.627 \\
836.0  Tear med menisc knee-cur  & 0.645 & 0.656 & 0.651 & 0.651 & 0.636 & 0.596 & 0.599 \\
599.0  Urin tract infection NOS  & 0.638 & 0.650 & 0.639 & 0.641 & 0.634 & 0.597 & 0.608 \\
715.90 Osteoarthros NOS-unspec  & 0.645 & 0.648 & 0.646 & 0.641 & 0.636 & 0.619 & 0.614 \\
790.21 Impaired fasting glucose  & 0.621 & 0.647 & 0.599 & 0.610 & 0.608 & 0.592 & 0.596 \\
716.90 Arthropathy NOS-unspec  & 0.590 & 0.614 & 0.627 & 0.643 & 0.638 & 0.626 & 0.621 \\
216.5  Benign neo skin trunk  & 0.626 & 0.620 & 0.631 & 0.643 & 0.621 & 0.589 & 0.604 \\
702.0  Actinic keratosis  & 0.639 & 0.642 & 0.632 & 0.635 & 0.637 & 0.580 & 0.582 \\
309.28 Adjust dis w anxiety/dep  & 0.633 & 0.642 & 0.628 & 0.617 & 0.619 & 0.592 & 0.602 \\
722.0  Cervical disc displacmnt  & 0.633 & 0.623 & 0.638 & 0.624 & 0.633 & 0.620 & 0.609 \\
780.53 Hypersom w slp apnea NOS  & 0.610 & 0.634 & 0.621 & 0.602 & 0.614 & 0.568 & 0.569 \\
703.0  Ingrowing nail  & 0.596 & 0.625 & 0.615 & 0.615 & 0.626 & 0.581 & 0.588 \\
401.1  Benign hypertension  & 0.608 & 0.624 & 0.614 & 0.610 & 0.611 & 0.597 & 0.594 \\
402.90 Hyp hrt dis NOS w/o hf  & 0.619 & 0.618 & 0.614 & 0.624 & 0.605 & 0.555 & 0.585 \\
726.5  Enthesopathy of hip  & 0.593 & 0.611 & 0.623 & 0.596 & 0.622 & 0.579 & 0.574 \\
366.16 Senile nuclear cataract  & 0.618 & 0.616 & 0.619 & 0.613 & 0.619 & 0.557 & 0.582 \\
702.19 Other sborheic keratosis  & 0.601 & 0.618 & 0.599 & 0.595 & 0.602 & 0.573 & 0.583 \\
739.1  Somat dysfunc cervic reg  & 0.607 & 0.613 & 0.617 & 0.605 & 0.606 & 0.572 & 0.584 \\
794.31 Abnorm electrocardiogram  & 0.600 & 0.615 & 0.616 & 0.608 & 0.607 & 0.588 & 0.592 \\
238.2  Unc behav neo skin  & 0.608 & 0.615 & 0.610 & 0.606 & 0.606 & 0.575 & 0.573 \\
272.4  Hyperlipidemia NEC/NOS  & 0.599 & 0.614 & 0.594 & 0.595 & 0.589 & 0.597 & 0.586 \\
723.4  Brachial neuritis NOS  & 0.605 & 0.604 & 0.610 & 0.600 & 0.603 & 0.571 & 0.584 \\
300.4  Dysthymic disorder  & 0.587 & 0.597 & 0.606 & 0.607 & 0.609 & 0.557 & 0.553 \\
715.09 General osteoarthrosis  & 0.607 & 0.608 & 0.608 & 0.608 & 0.605 & 0.583 & 0.584 \\
272.0  Pure hypercholesterolem  & 0.594 & 0.605 & 0.591 & 0.584 & 0.585 & 0.590 & 0.579 \\
995.3  Allergy, unspecified  & 0.572 & 0.571 & 0.596 & 0.596 & 0.600 & 0.574 & 0.577 \\
739.4  Somat dysfunc sacral reg  & 0.597 & 0.597 & 0.589 & 0.589 & 0.591 & 0.572 & 0.559 \\
372.14 Chr allrg conjunctiv NEC  & 0.575 & 0.596 & 0.592 & 0.584 & 0.577 & 0.552 & 0.574 \\
739.3  Somat dysfunc lumbar reg  & 0.585 & 0.595 & 0.575 & 0.585 & 0.582 & 0.565 & 0.564 \\
272.2  Mixed hyperlipidemia  & 0.573 & 0.582 & 0.557 & 0.562 & 0.557 & 0.571 & 0.566 \\
564.1  Irritable bowel syndrome  & 0.570 & 0.574 & 0.564 & 0.560 & 0.569 & 0.551 & 0.550 \\
785.1  Palpitations  & 0.565 & 0.570 & 0.570 & 0.565 & 0.565 & 0.555 & 0.551 \\

\hline
\end{longtable}
\end{center}

\end{document}